\documentclass[runningheads]{llncs}

\usepackage[mobile]{eccv}

\usepackage{eccvabbrv}

\usepackage{graphicx}
\usepackage{booktabs}
\usepackage[table]{colortbl} 
\usepackage{algorithm} 
\usepackage{algorithmicx} 
\usepackage{algpseudocode}
\usepackage{caption}
\usepackage{float}
\usepackage{placeins}
\usepackage{wrapfig}
\usepackage{xcolor}  
\usepackage{mdframed}  
\usepackage{amssymb}
\usepackage{tabularx}
\usepackage{tablefootnote}
\usepackage[accsupp]{axessibility}  
\usepackage{multirow}
\usepackage{listings}
\usepackage{footmisc}

\usepackage{color}
\usepackage{amssymb}

\usepackage{hyperref}

\usepackage{orcidlink}

\begin{document}

\title{CogView3: Finer and Faster Text-to-Image Generation via Relay Diffusion} 


\author{Wendi Zheng\inst{1}\protect\footnotemark[1]\protect\footnotemark[3] \and
Jiayan Teng\inst{1}\protect\footnotemark[1]\protect\footnotemark[3] \and
Zhuoyi Yang\inst{1}\protect\footnotemark[3] \and Weihan Wang\inst{1}\protect\footnotemark[3] \and\\ Jidong Chen\inst{1}\protect\footnotemark[3] \and Xiaotao Gu\inst{2} \and Yuxiao Dong\inst{1}\protect\footnotemark[2] \and Ming Ding\inst{2}\protect\footnotemark[2] \and Jie Tang\inst{1}\protect\footnotemark[2]}

\authorrunning{W.~Zheng et al.}

\institute{Tsinghua University\\
\email{\{zhengwd23@mails.,tengjy20@mails.,yuxiaod@,jietang@mail.\}tsinghua.edu.cn} \and Zhipu AI\\
\email{mingding.thu@gmail.com}}

\maketitle

\renewcommand{\thefootnote}{\fnsymbol{footnote}}
\footnotetext[1]{equal contribution}
\footnotetext[2]{corresponding authors}
\footnotetext[3]{work was done when interned in Zhipu AI.}

\vspace{-1.5em}
\begin{abstract}
 Recent advancements in text-to-image generative systems have been largely driven by diffusion models. However, single-stage text-to-image diffusion models still face challenges, in terms of computational efficiency and the refinement of image details. To tackle the issue, we propose CogView3, an innovative cascaded framework that enhances the performance of text-to-image diffusion. CogView3 is the first model implementing relay diffusion in the realm of text-to-image generation, executing the task by first creating low-resolution images and subsequently applying relay-based super-resolution. This methodology not only results in competitive text-to-image outputs but also greatly reduces both training and inference costs. Our experimental results demonstrate that CogView3 outperforms SDXL, the current state-of-the-art open-source text-to-image diffusion model, by 77.0\% in human evaluations, all while requiring only about 1/2 of the inference time. The distilled variant of CogView3 achieves comparable performance while only utilizing 1/10 of the inference time by SDXL.
\end{abstract}

\vspace{-1.6em}
\begin{figure}[h]
\begin{center}
{\includegraphics[width=1\linewidth]{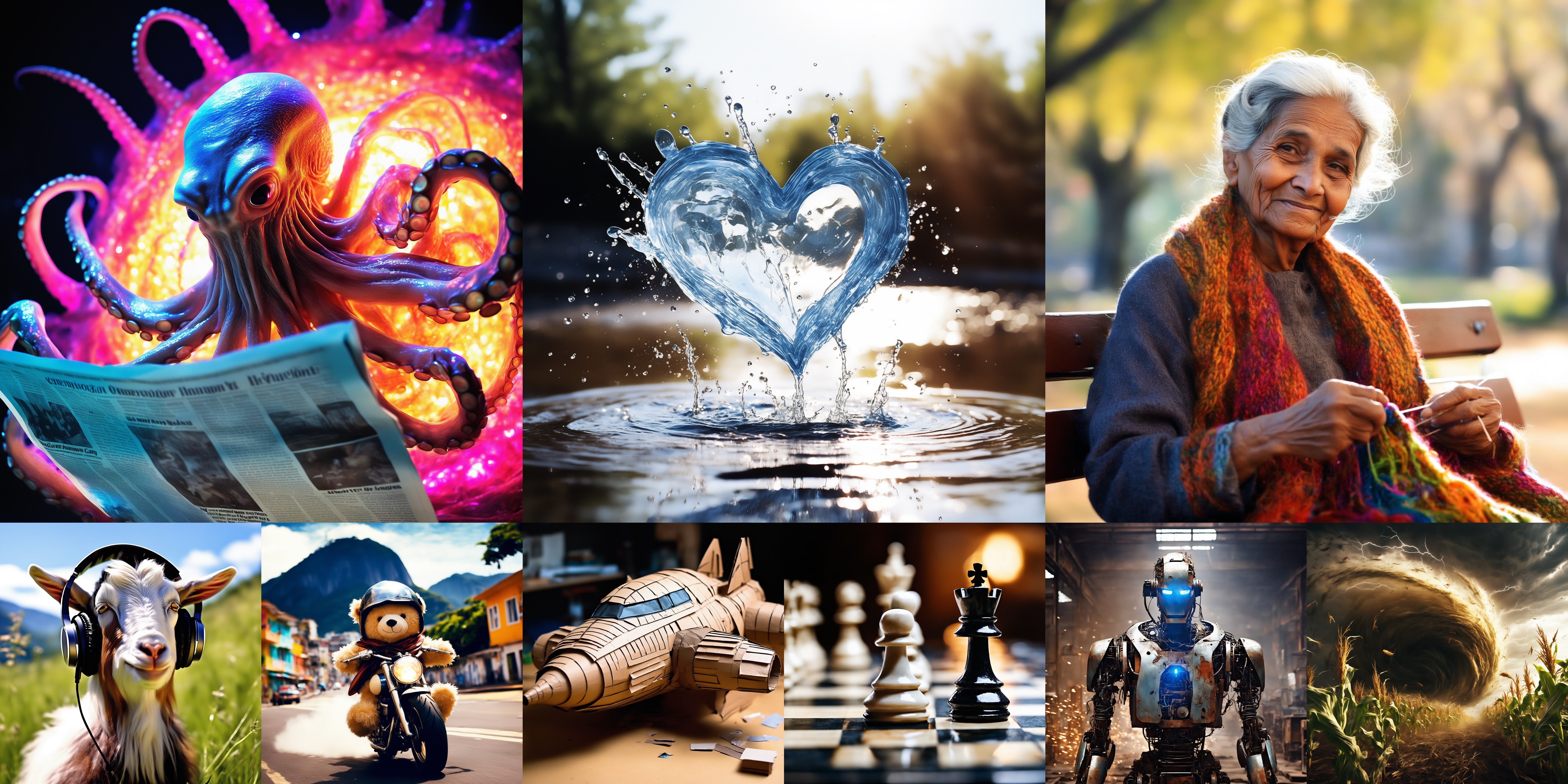}}
\end{center}
\vspace{-1em}
\caption{ Showcases of CogView3 generation of resolution $2048\times 2048$ \textbf{(top)} and $1024\times 1024$ \textbf{(bottom)}. All prompts are sampled from Partiprompts~\cite{yu2022scaling}. }
\label{fig:showcases}
\vspace{-1.5em}
\end{figure}

\vspace{-0.5em}
\section{Introduction}



Diffusion models have emerged as the mainstream framework in today's text-to-image generation systems~\cite{betker2023dalle3, podell2023sdxl, dai2023emu, saharia2022photorealistic, ramesh2022hierarchical}. In contrast to the paradigm of auto-regressive models~\cite{ding2021cogview, yu2022scaling, ramesh2021zero} and generative adversial networks~\cite{kang2023scaling}, the diffusion models conceptualize the task of image synthesis as a multi-step denoising process that starts from an isotropic Gaussian noise~\cite{ho2020denoising}. With the surge in the volume of training data and computation cost of neural networks, the framework of diffusion models has achieved effectiveness in the realm of visual generation, able to follow user instructions and generate images with commendable details.

Current state-of-the-art text-to-image diffusion models mostly operate in a single stage, conducting the diffusion process at high image resolutions, such as $1024\times 1024$~\cite{betker2023dalle3, podell2023sdxl, dai2023emu}. The direct modeling on high resolution images aggravates the inference costs since every denoising step is performed on the high resolution space. To address such an issue, Luo~\etal~\cite{luo2023latent} and Sauer~\etal~\cite{sauer2023adversarial} propose to distill diffusion models to significantly reduce the number of sampling steps. However, the generation quality tends to degrade noticeably during diffusion distillation, unless a GAN loss is introduced, which otherwise complicates the distillation and could lead to instability of training.

In this work, we propose CogView3, a novel text-to-image generation system that employs relay diffusion~\cite{teng2023relay}. Relay diffusion is a new cascaded diffusion framework, decomposing the process of generating high-resolution images into multiple stages. It first generates low-resolution images and subsequently performs relaying super-resolution generation. Unlike previous cascaded diffusion frameworks that condition every step of the super-resolution stage on low-resolution generations~\cite{ho2022cascaded, saharia2022photorealistic, ramesh2022hierarchical}, relaying super-resolution adds Gaussian noise to the low-resolution generations and starts diffusion from these noised images. This enables the super-resolution stage of relay diffusion to rectify unsatisfactory artifacts produced by the previous diffusion stage. In CogView3, we apply relay diffusion in the latent image space rather than at pixel level as the original version, by utilizing a simplified linear blurring schedule and a correspondingly formulated sampler. By the iterative implementation of the super-resolution stage, CogView3 is able to generate images with extremely high resolutions such as $2048\times 2048$.



Given that the cost of lower-resolution inference is quadratically smaller than that of higher-resolution, CogView3 can produce competitive generation results at significantly reduced inference costs by properly allocating sampling steps between the base and super-resolution stages. Our results of human evaluation show that CogView3 outperforms SDXL~\cite{podell2023sdxl} with a win rate of 77.0\%. Moreover, through progressive distillation of diffusion models, CogView3 is able to produce comparable results while utilizing only 1/10 of the time required for the inference of SDXL. Our contributions can be summarized as follows:

\vspace{-0.5em}
\begin{itemize}
    \item[$\bullet$] We propose CogView3, the first text-to-image system in the framework of relay diffusion. CogView3 is able to generate high quality images with extremely high resolutions such as $2048\times 2048$.
    \vspace{0.5em}
    \item[$\bullet$] Based on the relaying framework, CogView3 is able to produce competitive results at a significantly reduced time cost. CogView3 achieves a win rate of 77.0\% over SDXL with about 1/2 of the time during inference.
    \vspace{0.5em}
    \item[$\bullet$] We further explore the progressive distillation of CogView3, which is significantly facilitated by the relaying design. The distilled variant of CogView3 delivers comparable generation results while utilizing only 1/10 of the time required by SDXL.
\end{itemize}

\vspace{-1.5em}
\section{Background}

\subsection{Text-to-Image Diffusion Models}

Diffusion models, as defined by Ho~\etal~\cite{ho2020denoising}, establish a forward diffusion process that gradually adds Gaussian noise to corrupt real data $\Vec{x}_0$ as follows:
\begin{equation}
    q(\Vec{x}_t|\Vec{x}_{t-1})=\mathcal{N}(\Vec{x}_t; \sqrt{1-\beta_t}\Vec{x}_{t-1},\beta_t\mathbf{I}),\ t\in \{1,...,T\},
\end{equation}
where $\beta_t$ defines a noise schedule in control of diffusion progression. Conversely, the backward process generates images from pure Gaussian noise by step-by-step denoising, adhering a Markov chain. 

A neural network is trained at each time step to predict denoised results based on the current noised images. For text-to-image diffusion models, an additional text encoder encodes the text input, which is subsequently fed into the cross attention modules of the main network. The training process is implemented by optimizing the variational lower bound of the backward process, which is written as
\begin{equation}
    \mathbb{E}_{\Vec{x}_0\sim p_{data}}\mathbb{E}_{\mathbf{\Vec{\epsilon}}\sim\mathcal{N}(\mathbf{0}, \mathbf{I}),t}\Vert \mathcal{D}(\Vec{x}_0+\sigma_t \mathbf{\Vec{\epsilon}},t,c)-\Vec{x}_0\Vert^2,
\end{equation}
where $\sigma_t$ denotes the noise scale controlled by the noise schedule. $c$ denotes input conditions including the text embeddings.


Recent works~\cite{podell2023sdxl,betker2023dalle3} consistently apply diffusion models to the latent space, resulting in a substantial saving of both training and inference costs. They first use a pretrained autoencoder to compress the image $\Vec{x}$ into a latent representation $\Vec{z}$ with lower dimension, which is approximately recoverable by its decoder. The diffusion model learns to generate latent representations of images.

\vspace{-0.5em}
\subsection{Relay Diffusion Models}

Cascaded diffusion~\cite{ho2022cascaded,saharia2022photorealistic} refers to a multi-stage diffusion generation framework. It first generates low-resolution images using standard diffusion and subsequently performs super-resolution. The super-resolution stage of the original cascaded diffusion conditions on low-resolution samples $\Vec{x}^L$ at every diffusion step, by channel-wise concatenation of $\Vec{x}^L$ with noised diffusion states. Such conditioning necessitates augmentation techniques to bridge the gap in low-resolution input between real images and base stage generations.

As a new variant of cascaded diffusion, the super-resolution stage of relay diffusion~\cite{teng2023relay} instead starts diffusion from low-resolution images $\Vec{x}^L$ corrupted by Gaussian noise $\sigma_{T_r}\Vec{\epsilon}$, where $T_r$ denotes the starting point of the blurring schedule in the super-resolution stage. The forward process is formulated as:
\begin{equation}
    q(\Vec{x}_t | \Vec{x}_0) = \mathcal{N}(\Vec{x}_t | F(\Vec{x}_0, t), {\sigma_t}^2\mathbf{I}), \quad t\in\{0,...,T\},
\end{equation}
where $F(\cdot)$ is a pre-defined transition along time $t$ from high-resolution images $\Vec{x}=\Vec{x}_0$ to the upsampled low-resolution images $\Vec{x}^L$. The endpoint of $F$ is set as $F(\vec{x}_0, T_r)=\vec{x}^L$ to ensure a seamless transition. Conversely, the backward process of relaying super-resolution is a combination of denoising and deblurring.

This design allows relay diffusion to circumvent the need for intricate augmentation techniques on lower-resolution conditions $\vec{x}^L$, as $\vec{x}^L$ is only inputted at the initial sampling step of super-resolution stage and is already corrupted by Gaussian noise $\sigma_{T_r}\Vec{\epsilon}$. It also enables the super-resolution stage of relay diffusion to possibly rectify some unsatisfactory artifacts produced by the previous diffusion stage.

\vspace{-0.5em}
\subsection{Diffusion Distillation}


Knowledge distillation~\cite{hinton2015distilling} is a training process aiming to transfer a larger teacher model to the smaller student model. In the context of diffusion models, distillation has been explored as a means to reduce sampling steps thus saving computation costs of inference, while preventing significant degradation of the generation performance~\cite{salimans2022progressive, song2023consistency, luo2023latent, sauer2023adversarial}.

As one of the prominent paradigms in diffusion distillation, progressive distillation~\cite{salimans2022progressive} trains the student model to match every two steps of the teacher model with a single step in each training stage. This process is repeated, progressively halving the sampling steps. On the other hand, consistency models~\cite{song2023consistency, luo2023latent} propose a fine-tuning approach for existing diffusion models to project every diffusion step to the latest one to ensure step-wise consistency, which also reduces sampling steps of the model. While previous diffusion distillation methods mostly compromise on the quality of generation, adversial diffusion distillation~\cite{sauer2023adversarial} mitigates this by incorporating an additional GAN loss in the distillation. However, this makes the process of distillation more challenging due to the instability of GAN training.

\vspace{-0.5em}
\section{Method}

\subsection{Text Preprocessing}

\subsubsection{Image Recaption}



\begin{figure}[h]
\begin{center}
{\includegraphics[width=1\linewidth]{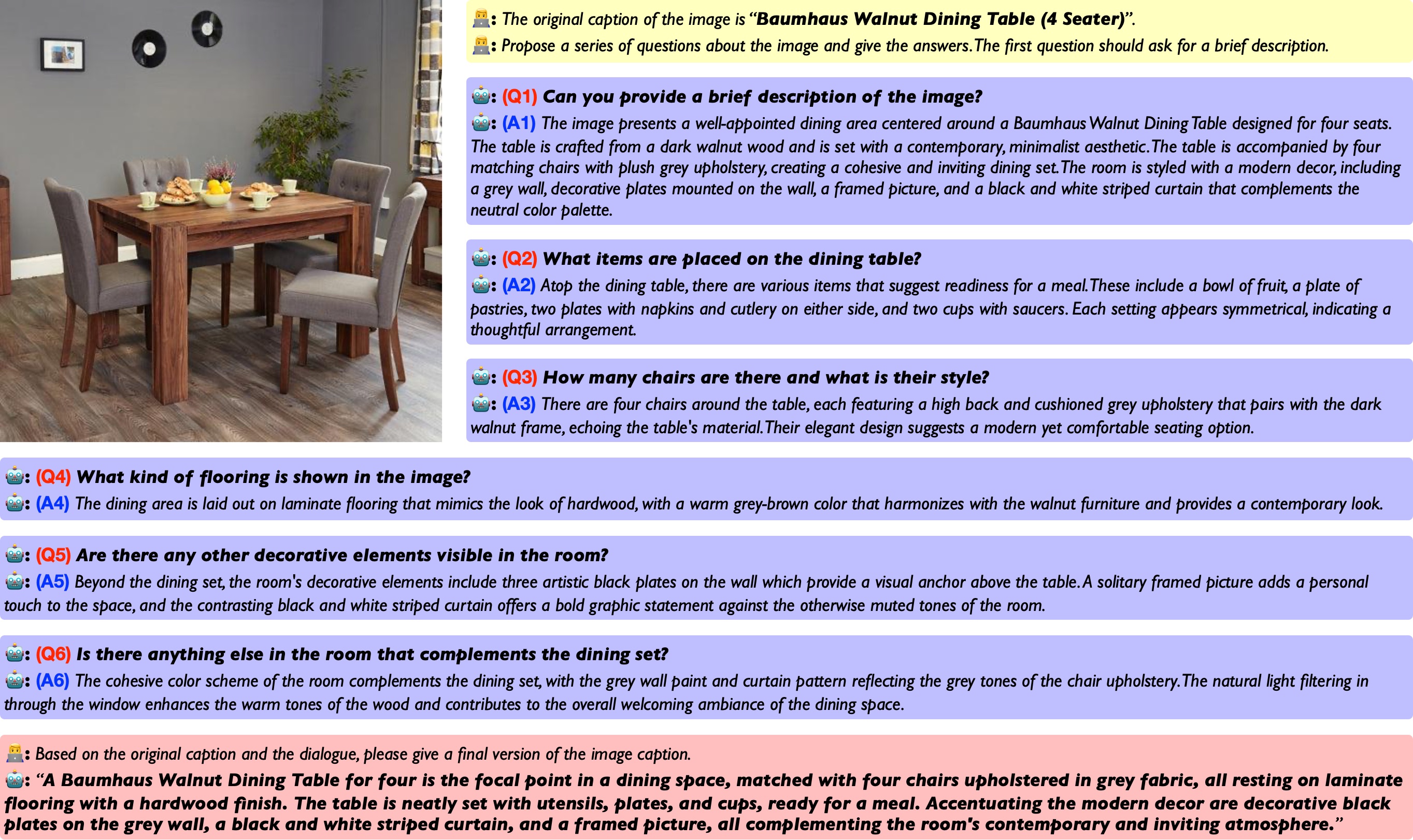}}
\end{center}
\vspace{-1em}
\caption{ An example of re-caption data collection from GPT-4V. }
\label{fig:recap}
\end{figure}

Following DALL-E-3~\cite{betker2023dalle3}, we develop an automatic pipeline to re-caption images from the training dataset. While DALL-E-3 derives instruction-tuning data of the re-caption model from human labelers, we extract triplets of \texttt{<image, old\_cap, new\_cap>} by automatically prompting GPT-4V~\cite{achiam2023gpt}, as shown in Figure~\ref{fig:recap}. Generally, we prompt GPT-4V to propose several questions about the content of the uploaded image. The first question is forced to be about a brief description. Finally, we instruct the model to combine the answers together with the original caption to build a new caption.

We collect approximately 70,000 recaption triplets with this paradigm and finetune CogVLM-17B~\cite{wang2023cogvlm} by these examples to obtain a recaption model. We finetune the model by a moderate degree, with batch size 256 and 1,500 steps to prevent model from severe overfitting. Eventually the model is utilized to re-caption the whole training dataset. The re-caption results provide comprehensive, graceful and detailed descriptions of images, in contrast to the original short and less relevant captions from the dataset. The prefix statement we use to prompt GPT-4V and the template we use in fine-tuning the recaption model are both provided in Appendix~\ref{sec:app_text}.


\vspace{-0.5em}

\vspace{-2.5em}
\subsubsection{Prompt Expansion}
\label{sec:text_process}

On account that CogView3 is trained on datasets with comprehensive re-captions while users of text-to-image generation systems may tend to provide brief prompts lacking descriptive information, this introduces an explicit misalignment between model training and inference~\cite{betker2023dalle3}. Therefore, we also explore to expand user prompts before sampling with the diffusion models. We prompt language models to expand user prompts into comprehensive descriptions, while encouraging the model generation to preserve the original intention from users. With human evaluation, we find results of the expanded prompts to achieve higher preference. We provide the template and showcases of our prompt expansion in Appendix~\ref{sec:app_text}.

\vspace{-1em}
\subsection{Model Formulation}

\vspace{-1em}
\begin{figure}[h]
\begin{center}
{\includegraphics[width=1\linewidth]{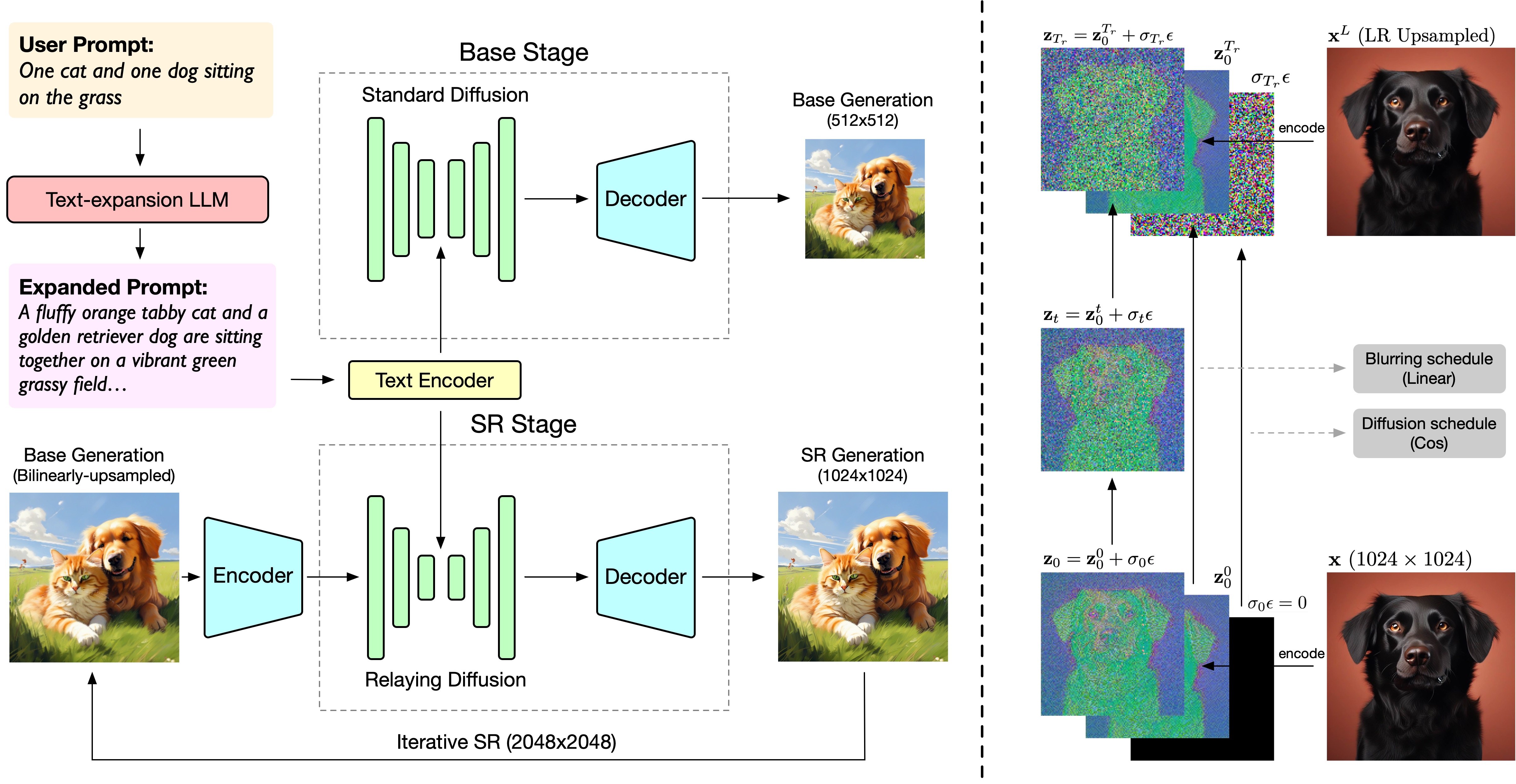}}
\end{center}
\vspace{-1em}
\caption{ \textbf{(left)} The pipeline of CogView3. User prompts are rewritten by a text-expansion language model. The base stage model generates $512\times 512$ images, and the second stage subsequently performs relaying super-resolution. \textbf{(right)} Formulation of relaying super-resolution in the latent space.}
\label{fig:pipeline}
\vspace{-1.2em}
\end{figure}


\subsubsection{Model Framework}

The backbone of CogView3 is a 3-billion parameter text-to-image diffusion model with a 3-stage UNet architecture. The model operates in the latent image space, which is $8\times$ compressed from the pixel space by a variational KL-regularized autoencoder. We employ the pretrained T5-XXL~\cite{raffel2020exploring} encoder as the text encoder to improve model's capacity for text understanding and instruction following, which is frozen during the training of the diffusion model. To ensure alignment between training and inference, user prompts are first rewritten by language models as mentioned in the previous section. We set the input token length for the text encoder as 225 to facilitate the implementation of the expanded prompts.


As shown in Figure~\ref{fig:pipeline}(left), CogView3 is implemented as a 2-stage relay diffusion. The base stage of CogView3 is a diffusion model that generates images at a resolution of $512\times 512$. The second stage model performs $2\times$ super-resolution, generating $1024\times 1024$ images from $512\times 512$ inputs. It is noteworthy that the super-resolution stage can be directly transferred to higher resolutions and iteratively applied, enabling the final outputs to reach higher resolutions such as $2048\times 2048$, as cases illustrated from the top line of Figure~\ref{fig:showcases}.


\subsubsection{Training Pipeline}


We use Laion-2B~\cite{schuhmann2022laion} as our basic source of the training dataset, after removing images with politically-sensitive, pornographic or violent contents to ensure appropriateness and quality of the training data. The filtering process is executed by a pre-defined list of sub-strings to block a group of source links associated with unwanted images. In correspondence with Betker~\etal~\cite{betker2023dalle3}, we replace 95\% of the original data captions with the newly-produced captions.

Similar to the training approach used in SDXL~\cite{podell2023sdxl}, we train Cogview3 progressively to develop multiple stages of models. This greatly reduced the overall training cost. Owing to such a training setting, the different stages of CogView3 share a same model architecture.



The base stage of CogView3 is trained on the image resolution of $256\times 256$ for 600,000 steps with batch size 2048 and continued to be trained on $512\times 512$ for 200,000 steps with batch size 2048. We finetune the pretrained $512\times 512$ model on a highly aesthetic internal dataset for 10,000 steps with batch size 1024, to achieve the released version of the base stage model. To train the super-resolution stage of CogView3, we train on the basis of the pretrained $512\times 512$ model on $1024\times 1024$ resolution for 100,000 steps with batch size 1024, followed by a 20,000 steps of finetuning with the loss objective of relaying super-resolution to achieve the final version.

\vspace{-1em}
\subsection{Relaying Super-resolution}

\subsubsection{Latent Relay Diffusion}

The second stage of CogView3 performs super-resolution by relaying, starting diffusion from the results of base stage generation. While the original relay diffusion handles the task of image generation in the pixel level~\cite{teng2023relay}, we implement relay diffusion in the latent space and utilize a simple linear transformation instead of the original patch-wise blurring. The formulation of latent relay diffusion is illustrated by Figure~\ref{fig:pipeline}(right). Given an image $\Vec{x}_0$ and its low-resolution version $\Vec{x}^L=\text{Downsample}(\Vec{x}_0)$, they are first transformed into latent space by the autoencoder as $\Vec{z}_0=\mathcal{E}(\Vec{x}_0),\ \Vec{z}^L=\mathcal{E}(\Vec{x}^L)$. Then the linear blurring transformation is defined as:


\begin{equation}\label{eq:zt_define}
    \Vec{z}_0^t=\mathcal{F}(\Vec{z}_0, t)=\frac{T_r-t}{T_r}\Vec{z}_0+\frac{t}{T_r}\Vec{z}^L,
\end{equation}
where $T_r$ denotes the starting point set for relaying super-resolution and $\Vec{z}_0^{T_r}$ matches exactly with $\Vec{z}^L$. The forward process of the latent relay diffusion is then written as:

\begin{equation}\label{eq:relay_forward}
    q(\Vec{z}_t|\Vec{z}_0)=\mathcal{N}(\Vec{z}_t|\Vec{z}_0^t, \sigma^2_t\mathbf{I}),\ t\in\{1,...,T_r\}.
\end{equation}
The training objective is accordingly formulated as:

\begin{equation}
    \mathbb{E}_{\Vec{x}_0\sim p_{data}}\mathbb{E}_{\Vec{\epsilon}\sim \mathcal{N}(\mathbf{0},\mathbf{I}),t\in\{0,...,T_r\}}\Vert 
\mathcal{D}(\Vec{z}_0^t+\sigma_t\Vec{\epsilon},t,c_{text})-\Vec{z}_0 \Vert^2,
\end{equation}
where $\mathcal{D}$ denotes the UNet denoiser function and $c_{text}$ denotes the input text condition.

\subsubsection{Sampler Formulation}

Next we introduce the sampler designed for the relaying super-resolution.
Given samples $X^L$ generated in the base stage, we bilinearly upsample $X^L$ into $\Vec{x}^L$. The starting point of relay diffusion is defined as $\Vec{z}_{T_r}=\Vec{z}_0^{T_r}+\sigma_{T_r}\Vec{\epsilon}$, where $\Vec{\epsilon}$ denotes a unit isotropic Gaussian noise and $\Vec{z}_0^{T_r}=\mathcal{E}(\Vec{x}^L)$ is the latent representation of the bilinearly-upsampled base-stage generation.
Corresponding to the forward process of relaying super-resolution formulated in Equation~\ref{eq:relay_forward}, the backward process is defined in the DDIM~\cite{song2020denoising} paradigm:
\begin{equation}~\label{eq:back_process}
    q(\Vec{z}_{t-1} | \Vec{z}_t,\Vec{z}_0) = \mathcal{N}(\Vec{z}_{t-1} | a_t\Vec{z}_t + b_t\Vec{z}_0 + c_t\Vec{z}_0^t, \delta_t^2\mathbf{I}),
\end{equation}
where $a_t=\sqrt{\sigma_{t-1}^2-\delta_t^2}/\sigma_t$, $b_t=1/t$, $c_t=(t-1)/t - a_t$, $\Vec{z}_0^t$ is defined in Equation~\ref{eq:zt_define} and $\delta_t$ represents the random degree of the sampler. In practice, we simply set $\delta_t$ as 0 to be an ODE sampler.
The procedure is shown in Algorithm~\ref{alg:sample}. A detailed proof of the consistency between the sampler and the formulation of latent relay diffusion is shown in Appendix~\ref{sec:app_sampler}.



\vspace{-1em}
\begin{algorithm}
\caption{latent relay sampler}
\begin{algorithmic}
\State {{\bf Given} $\Vec{x}^L, \Vec{z}_0^{T_r}=\mathcal{E}(\Vec{x}^L)$}
\State $\Vec{z}_{T_r}=\Vec{z}_0^{T_r}+\sigma_{T_r}\Vec{\epsilon}$
    \textcolor{gray}{\Comment{transform into the latent space and add noise for relaying}}
\For{$t \in \{T_r, \dots, 1\}$}
    \State $\Tilde{\Vec{z}}_0=\mathcal{D}(\Vec{z}_t,t,c_{text})$
        \textcolor{gray}{\Comment{predict $\Vec{z}_0$}}
    \vspace{1mm}
    \State $\Vec{z}_0^{t-1}=\Vec{z}_0^t+(\Tilde{\Vec{z}}_0-\Vec{z}_0^t)/t$
        \textcolor{gray}{\Comment{linear blurring transition}}
    \vspace{1mm}
    \State $a_t=\sigma_{t-1}/\sigma_t, b_t=1/t, c_t=(t-1)/t-a_t$
        \textcolor{gray}{\Comment{coefficient of each item}}
    \State {$\Vec{z}_{t-1}=a_t\Vec{z}_t+b_t\Tilde{\Vec{z}}_0+c_t\Vec{z}_0^t$} 
        \textcolor{gray}{\Comment{single sampling step}}
\EndFor
\State $\Vec{x}_0=\text{Decode}(\Vec{z}_0)$
\end{algorithmic}
\label{alg:sample}
\end{algorithm}
\vspace{-3em}

\subsection{Distillation of Relay Diffusion}

We combine the method of progressive distillation~\cite{meng2023distillation} and the framework of relay diffusion to achieve the distilled version of CogView3. While the base stage of CogView3 performs standard diffusion, the distillation procedure follows the original implementation.

For the super-resolution stage, we merge the blurring schedule into the diffusion distillation training, progressively halving sampling steps by matching two steps from the latent relaying sampler of the teacher model with one step of the student model. The teacher steps are formulated as:

\begin{equation}
\begin{aligned}
    \Vec{z}_{t-1} &=a_t\Vec{z}_t+b_t\Tilde{\Vec{z}}_0(\Vec{z}_t,t)_{teacher}+c_t\Vec{z}_0^t, \\
    \Vec{z}_{t-2} &=a_{t-1}\Vec{z}_{t-1}+b_{t-1}\Tilde{\Vec{z}}_0(\Vec{z}_{t-1},t-1)_{teacher}+c_{t-1}\Vec{z}_0^{t-1}, \\
\end{aligned}
\end{equation}

where $(a_k,b_k,c_k),\ k\in\{0,...,T_r\}$ refers to the item coefficients defined in Algorithm~\ref{alg:sample}. One step of the student model is defined as:

\begin{equation}
    \Vec{\hat{z}}_{t-2} = \frac{\sigma_{t-2}}{\sigma_t}\Vec{z}_t + \frac{\Tilde{\Vec{z}}_0(\Vec{z}_t,t)_{student}}{t} + (\frac{t-2}{t}-\frac{\sigma_{t-2}}{\sigma_t})\Vec{z}_0^t.
\end{equation}

The training objective is defined as the mean square error between $\Vec{\hat{z}}_{t-2}$ and $\Vec{z}_{t-2}$. Following Meng~\etal~\cite{meng2023distillation}, we incorporate the property of the classifier-free guidance (CFG)~\cite{ho2022classifier} strength $w$ into the diffusion model in the meantime of distillation by adding learnable projection embeddings of $w$ into timestep embeddings. Instead of using an independent stage for the adaptation, we implement the incorporation at the first round of the distillation and directly condition on $w$ at subsequent rounds.

The inference costs of the low-resolution base stage are quadratically lower than the high-resolution counterparts, while it ought to be called from a complete diffusion schedule. On the other hand, the super-resolution stage starts diffusion at an intermediate point of the diffusion schedule. This greatly eases the task and reduces the potential error that could be made by diffusion distillation. Therefore, we are able to distribute final sampling steps for relaying distillation as 8 steps for the base stage and 2 steps for the super-resolution stage, or even reduce to 4 steps and 1 step respectively, which achieves both greatly-reduced inference costs and mostly-retained generation quality.

\section{Experiments}

\subsection{Experimental Setting}


We implement a comprehensive evaluation process to demonstrate the performance of CogView3. With an overall diffusion schedule of 1000 time steps, we set the starting point of the relaying super-resolution at 500, a decision informed by a brief ablation study detailed in Section~\ref{sec:start_abl}. To generate images for comparison,  we sample 50 steps by the base stage of CogView3 and 10 steps by the super-resolution stage, both utilizing a classifier-free guidance~\cite{ho2022classifier} of 7.5, unless specified otherwise. The comparison is all conducted at the image resolution of $1024\times 1024$.

\subsubsection{Dataset}


We choose a combination of image-text pair datasets and collections of prompts for comparative analysis. Among these, MS-COCO~\cite{lin2014microsoft} is a widely applied dataset for evaluating the quality of text-to-image generation. We randomly pick a subset of 5000 image-text pairs from MS-COCO, named as COCO-5k. We also incorporate DrawBench~\cite{saharia2022photorealistic} and PartiPrompts~\cite{yu2022scaling}, two well-known sets of prompts for text-to-image evaluation. DrawBench comprises 200 challenging prompts that assess both the quality of generated samples and the alignment between images and text. In contrast, PartiPrompts contains 1632 text prompts and provides a comprehensive evaluation critique.

\subsubsection{Baselines}


In our evaluation, we employ state-of-the-art open-source text-to-image models, specifically SDXL~\cite{podell2023sdxl} and Stable Cascade~\cite{pernias2023wuerstchen} as our baselines. SDXL is a single-stage latent diffusion model capable of generating images at and near a resolution of $1024\times 1024$. On the other hand, Stable Cascade implements a cascaded pipeline, generating $16\times 24\times 24$ priors at first and subsequently conditioning on the priors to produce images at a resolution of $1024\times 1024$. We sample SDXL for 50 steps and Stable Cascade for 20 and 10 steps respectively for its two stages. In all instances, we adhere to their recommended configurations of the classifier-free guidance.

\vspace{-1em}
\subsubsection{Evaluation Metrics}

We use Aesthetic Score (Aes)~\cite{schuhmann2022laion} to evaluate the image quality of generated samples. We also adopt Human Preference Score v2 (HPS v2)~\cite{wu2023human} and ImageReward~\cite{xu2024imagereward} to evaluate text-image alignment and human preference. Aes is obtained by an aesthetic score predictor trained from LAION datasets, neglecting alignment of prompts and images. HPS v2 and ImageReward are both used to predict human preference for images, including evaluation of text-image alignment, human aesthetic, etc. Besides machine evaluation, we also conduct human evaluation to further assess the performance of models, covering image quality and semantic accuracy.

\vspace{-0.5em}
\subsection{Results of Machine Evaluation}\label{sec:machine_eval}



Table~\ref{tbl:result} shows results of machine metrics on DrawBench and Partiprompts. While CogView3 has the lowest inference cost, it outperforms SDXL and Stable Cascade in most of the comparisons except for a slight setback to Stable Cascade on the ImageReward of PartiPrompts. Similar results are observed from comparisons on COCO-5k, as shown in Table~\ref{tbl:result_coco}. The distilled version of CogView3 takes an extremely low inference time of 1.47s but still achieves a comparable performance. The results of the distilled variant of CogView3 significantly outperform the previous distillation paradigm of latent consistency model~\cite{luo2023latent} on SDXL, as illustrated in the table.

\vspace{-0.5em}
\begin{table}[H]
    \vspace{-1mm}
    \begin{center}
    \resizebox{\textwidth}{!}{
    \begin{small}
    \begin{tabular}{lcc|ccc|ccc}
    \toprule
    \multicolumn{1}{l}{\multirow{2}{*}{\large{Model}}} & \multicolumn{1}{c}{\multirow{2}{*}{Steps}} & \multicolumn{1}{c|}{\multirow{2}{*}{Time Cost}} & \multicolumn{3}{c|}{\textbf{DrawBench}} & \multicolumn{3}{c}{\textbf{PartiPrompts}} \\
    \cline{4-9}
     & & & Aes$\uparrow$ & HPS v2$\uparrow$ & ImageReward$\uparrow$ & Aes$\uparrow$ & HPS v2$\uparrow$ & ImageReward$\uparrow$ \\
    \midrule
    SDXL~\cite{podell2023sdxl}  &50 & 19.67s & 5.54 & \underline{0.288} & 0.676 & 5.78 & 0.287 & 0.915 \\
    StableCascade~\cite{pernias2023wuerstchen} &20+10 & 10.83s & 5.88 & 0.285 & 0.677 & 5.93 & 0.285 & \bf{1.029} \\
    \textbf{CogView3}  &50+10 & \bf{10.33s} & \bf{5.97} & \bf{0.290} & \bf{0.847} & \bf{6.15} & \bf{0.290} & \underline{1.025} \\
    \midrule
    LCM-SDXL~\cite{luo2023latent}  &4 & 2.06s & 5.45 & 0.279 & 0.394 & 5.59 & 0.280 & 0.689 \\
    \textbf{CogView3-distill}  &4+1 & \bf{1.47s} & 5.87 & \underline{0.288} & \underline{0.731} & 6.12 & 0.287 & 0.968  \\
    \textbf{CogView3-distill}  &8+2 & 1.96s & \underline{5.90} & 0.285 & 0.655 & \underline{6.13} & \underline{0.288}  & 0.963  \\
    \bottomrule
    \end{tabular}
    \end{small}
    }
    \end{center}
    \caption{Results of machine metrics on DrawBench and PartiPrompts. All samples are generated on $1024\times 1024$. The time cost is measured with a batch size of 4. } %
    \label{tbl:result}
    \vspace{-2.5em}
\end{table}

\begin{table}[H]
    \vspace{-6mm}
    \begin{center}
    \begin{small}
    \begin{tabular}{lcccccc}
    \toprule
    \multicolumn{7}{l}{\bf{COCO-5k }} \\
    \toprule
    Model                       & Steps & Time Cost & FID$\downarrow$   & Aes$\uparrow$   & HPS v2$\uparrow$  & ImageReward$\uparrow$ \\
    \midrule
    SDXL~\cite{podell2023sdxl}  & 50    & 19.67s    & \bf{26.29}  & 5.63                 & 0.291                 & 0.820  \\
    StableCascade~\cite{pernias2023wuerstchen}               & 20+10 & 10.83s    & 36.59       & 5.89                 & 0.283                 & 0.734  \\
    \textbf{CogView3}           & 50+10 & \bf{10.33s}    & 31.63       & \bf{6.01}            & \bf{0.294}            & \bf{0.967}  \\
    \midrule
    LCM-SDXL~\cite{luo2023latent}                    & 4     & 2.06s    & \underline{27.16}  & 5.39                 & 0.281                 & 0.566  \\
    \textbf{CogView3-distill}   & 4+1   & \bf{1.47s}    & 34.03       & 5.99                 & 0.292                 & 0.920  \\
    \textbf{CogView3-distill}   & 8+2   & 1.96s    & 35.53       & \underline{6.00}     & \underline{0.293}     & \underline{0.921}  \\
    \bottomrule
    \end{tabular}
    \end{small}
    \end{center}
    \caption{Results of machine metrics on COCO-5k. All samples are generated on $1024\times 1024$. The time cost is measured with a batch size of 4.} %
    \label{tbl:result_coco}
\end{table}
\vspace{-2em}

The comparison results demonstrate the performance of CogView3 for generating images of improved quality and fidelity with a remarkably reduced cost. The distillation of CogView3 succeeds in preserving most of the generation quality while reduces the sampling time to an extreme extent. We largely attribute the aforementioned comparison results to the relaying property of CogView3. In the following section, we will further demonstrate the performance of CogView3 with human evaluation.

\vspace{-0.6em}
\subsection{Results of Human Evaluation}

\vspace{-2em}
\begin{figure}[h]
\begin{center}
{\includegraphics[width=1\linewidth]{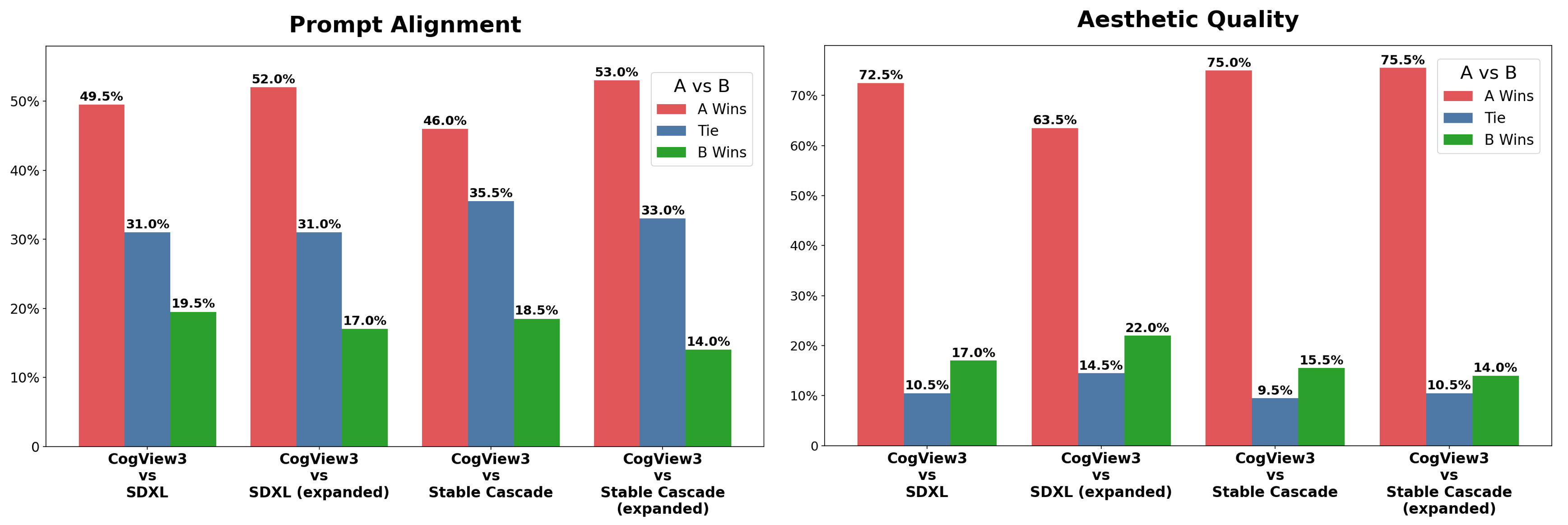}}
\end{center}
\vspace{-1.5em}
\caption{Results of human evaluation on DrawBench generation. \textbf{(left)} Comparison results about prompt alignment, \textbf{(right)} comparison results about aesthetic quality. ``(expanded)'' indicates that prompts used for generation is text-expanded.}
\label{fig:DB_1}
\end{figure}

\vspace{-3.5em}
\begin{figure}[h]
\begin{center}
{\includegraphics[width=1\linewidth]{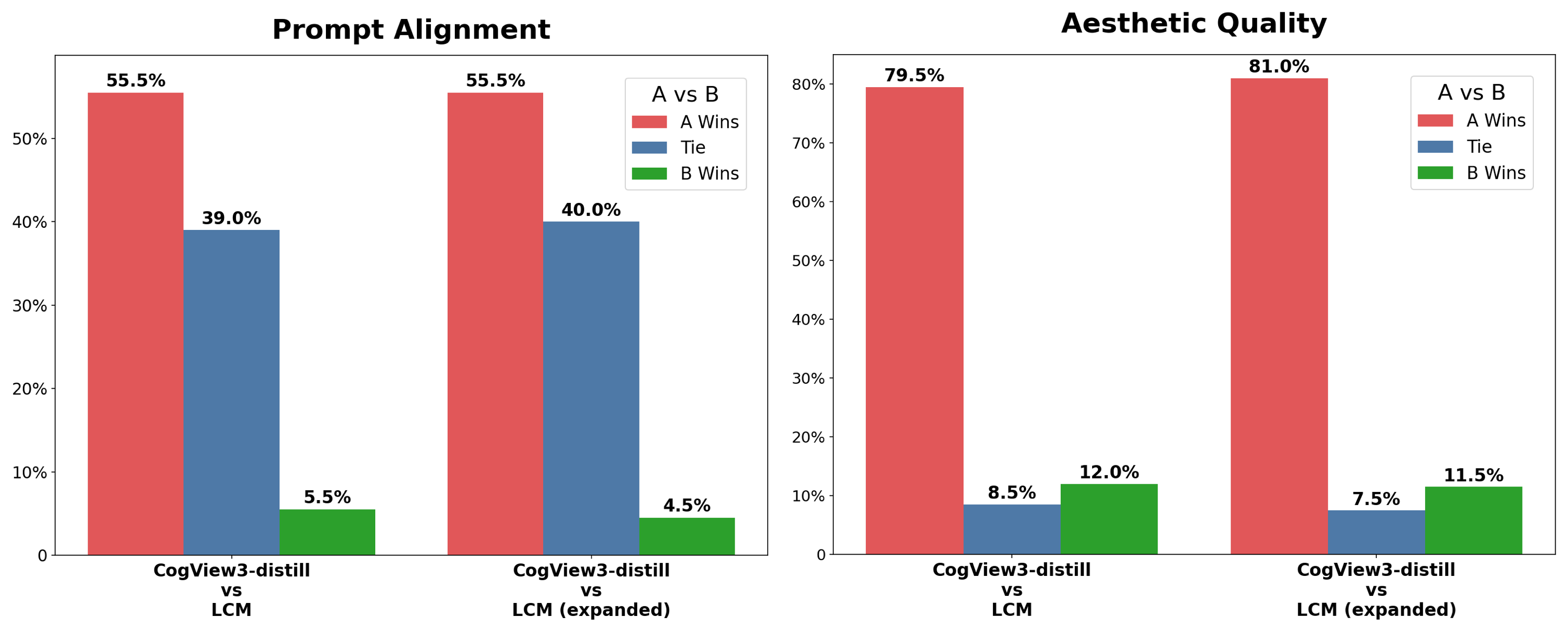}}
\end{center}
\vspace{-1.5em}
\caption{Results of human evaluation on Drawbench generation for distilled models. \textbf{(left)} Comparison results about prompt alignment, \textbf{(right)} comparison results about aesthetic quality. ``(expanded)'' indicates that prompts used for generation is text-expanded. We sample 8+2 steps for CogView3-distill and 4 steps for LCM-SDXL.}
\label{fig:DB_2}
\vspace{-1em}
\end{figure}



We conduct human evaluation for CogView3 by having annotators perform pairwise comparisons. The human annotators are asked to provide results of win, lose or tie based on the prompt alignment and aesthetic quality of the generation. We use DrawBench~\cite{saharia2022photorealistic} as the evaluation benchmark. For the generation of CogView3, we first expand the prompts from DrawBench to detailed descriptions as explained in Section~\ref{sec:text_process}, using the expanded prompts as the input of models. For a comprehensive evaluation, we compare CogView3 generation with SDXL and Stable Cascade by both the original prompts and the expanded prompts. 


As shown in Figure~\ref{fig:DB_1}, CogView3 significantly outperforms SDXL and Stable Cascade in terms of both prompt alignment and aesthetic quality, achieving average win rates of 77.0\% and 78.1\% respectively. Similar results are observed on comparison with SDXL and Stable Cascade generation by the expanded prompts, where CogView3 achieves average win rates of 74.8\% and 82.1\% respectively.


To evaluate the distillation, we compare the distilled CogView3 with SDXL distilled in the framework of latent consistency model~\cite{luo2023latent}. As shown in Figure~\ref{fig:DB_2}, the performance of the distilled CogView3 significantly surpasses that of LCM-distilled SDXL, which is consistent with the results from Section~\ref{sec:machine_eval}.



\subsection{Additional Ablations}

\subsubsection{Starting Points for Relaying Super-resolution}\label{sec:start_abl}

\begin{figure}[h]
\begin{center}
{\includegraphics[width=1\linewidth]{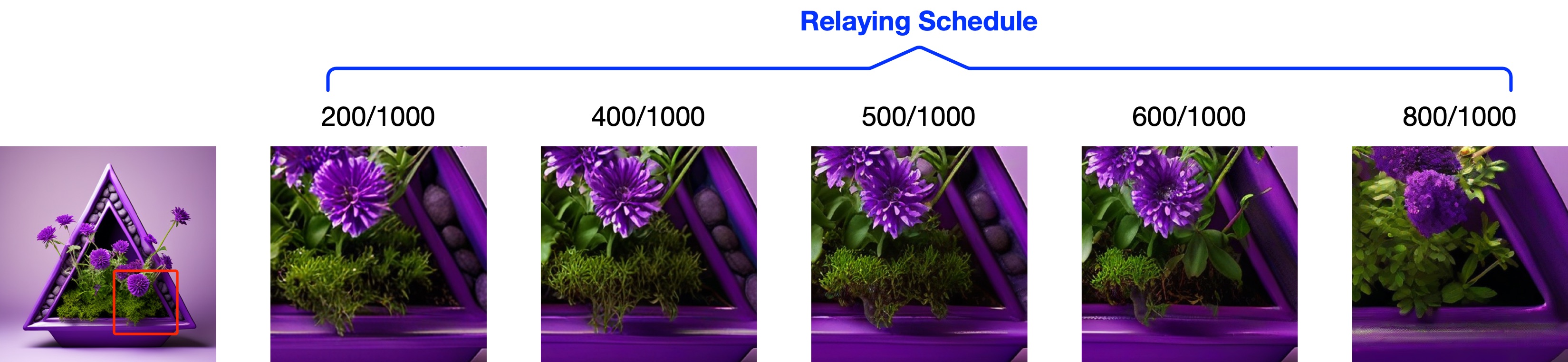}}
\end{center}
\caption{Comparison of results from super-resolution stages with different relaying starting points. Sampling steps are all set $\sim$10 by controlling the number of steps from the complete diffusion schedule.}
\label{fig:start_compare}
\end{figure}

We ablate the selection of starting point for relaying super-resolution as shown in Table~\ref{tbl:result_start_abl}, finding that a midway point achieves the best results. The comparison is also illustrated with a qualitative case in Figure~\ref{fig:start_compare}. An early starting point tends to produce blurring contents, as shown by the flower and grass in case of \texttt{200/1000}, while in contrast, a late starting point introduces artifacts, as shown by the flower and edge in case of \texttt{800/1000}, suggesting a midway point to be the best choice. Based on the results of comparison, we choose 500 as our finalized starting point. 

\vspace{1.5em}
\begin{table}[h]
    \vspace{-1em}
    \begin{center}
    \begin{small}
    \begin{tabularx}{\textwidth}{lXXXXX} 
    \toprule
     Starting Point             & 200/1000   & 400/1000   & 500/1000   & 600/1000   & 800/1000 \\
    \midrule
     HPS v2 $\uparrow$           & 0.288 & 0.289 & \bf{0.290} & 0.289 & 0.286 \\
     ImageReward $\uparrow$      & 0.829 & 0.835 & \bf{0.847} & 0.836 & 0.812 \\
    \bottomrule
    \end{tabularx}
    \end{small}
    \end{center}
    \caption{Ablation of starting points on DrawBench. } %
    \label{tbl:result_start_abl}
    \vspace{-2em}
\end{table}

\vspace{-0.5em}

\subsubsection{Alignment Improvement with Text Expansion}



\begin{wrapfigure}{r}{.5\textwidth}
    \includegraphics[width=\linewidth]{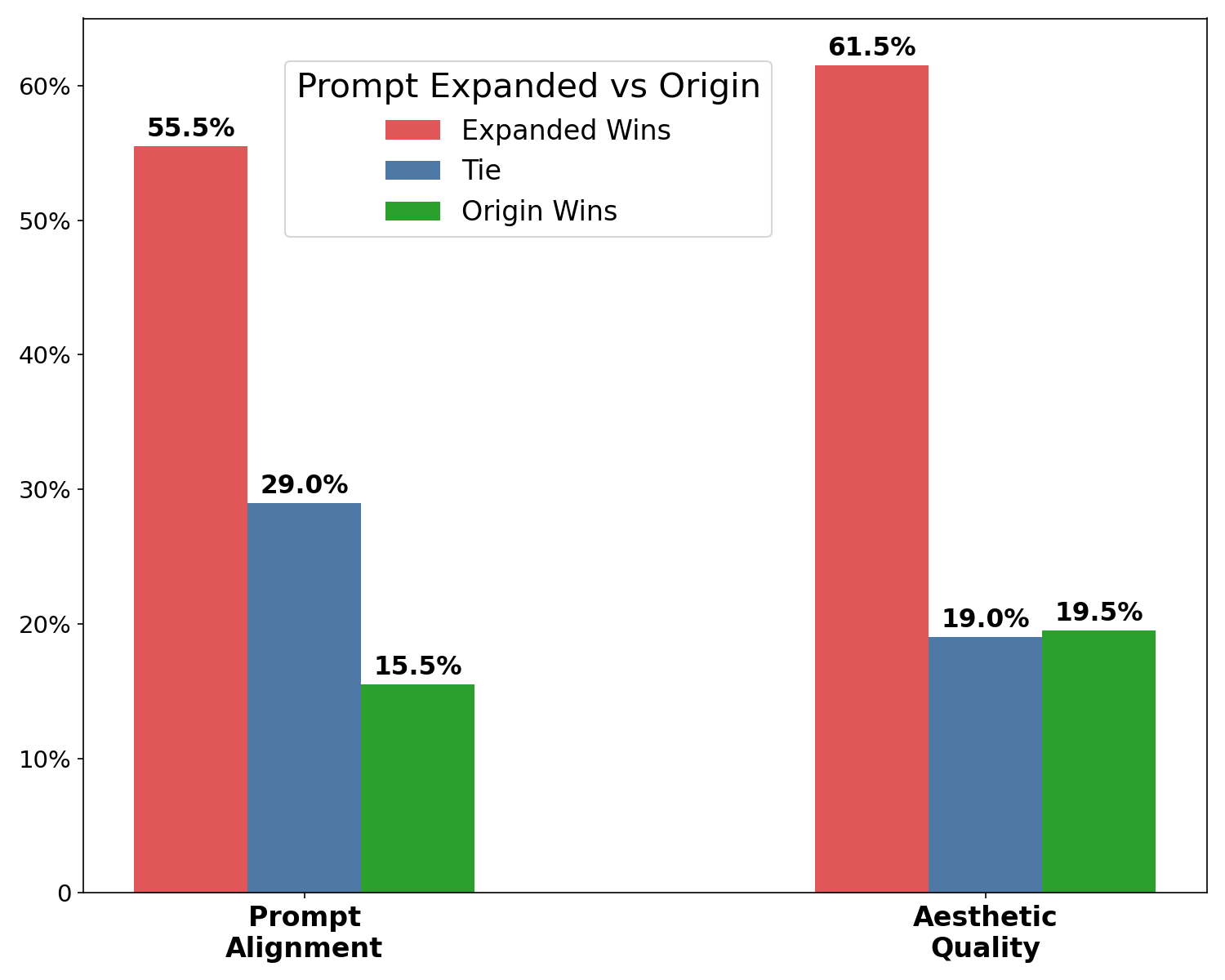}
    \caption{Human evaluation results of CogView3 before and after prompt expansion on DrawBench.}
    \vspace{-2em}
    \label{fig:DB_3}
\end{wrapfigure}

\begin{figure}[h]
\begin{center}
{\includegraphics[width=1\linewidth]{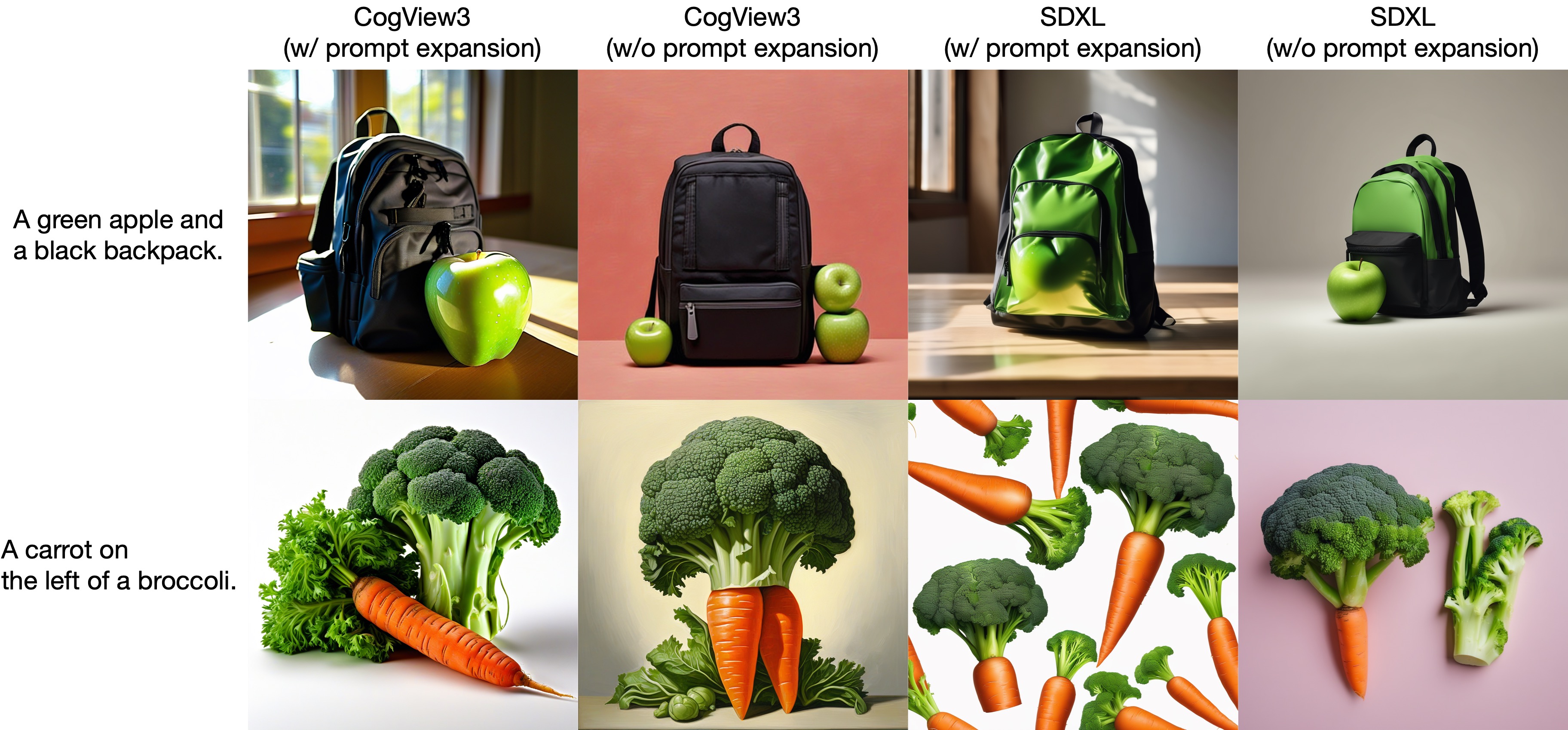}}
\end{center}
\caption{Comparison of the effect of prompt expansion for CogView3 and SDXL.}
\label{fig:prompt_compare}
\end{figure}

While prompt expansion hardly brings an improvement for the generation of SDXL and Stable Cascade, we highlight its significance for the performance of CogView3. Figure~\ref{fig:DB_3} shows the results of comparison with and without prompt expansion, explicitly demonstrating that prompt expansion significantly enhances the ability of prompt instruction following for CogView3. Figure~\ref{fig:prompt_compare} shows qualitative comparison between before and after the prompt expansion. The expanded prompts provide more comprehensive and in-distribution descriptions for model generation, largely improving the accuracy of instruction following for CogView3. Similar improvement is not observed on the generation of SDXL. The probable reason may be that SDXL is trained on original captions and only has an input window of 77 tokens, which leads to frequent truncation of the expanded prompts. This corroborates the statement in Section~\ref{sec:text_process} that prompt expansion helps bridge the gap between model inference and training with re-captioned data.

\vspace{1em}

\subsubsection{Methods of Iterative Super-Resolution}

\begin{figure}[h]
\begin{center}
{\includegraphics[width=1\linewidth]{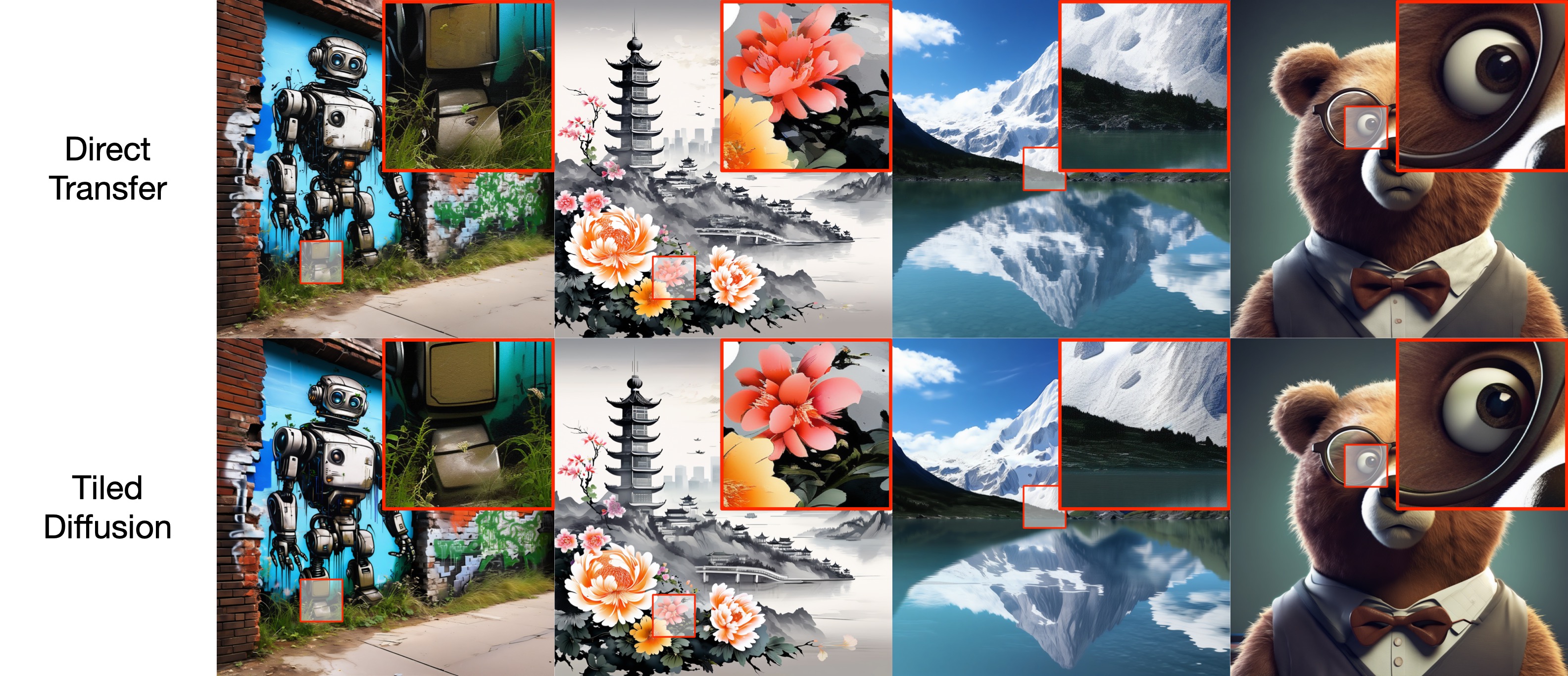}}
\end{center}
\caption{Comparison of direct higher super-resolution and tiled diffusion on $2048\times 2048$. We choose Mixture of Diffusers~\cite{jimenez2023mixture} in view of its superior quality of integration. Original prompts are utilized for the inference of all blocks.}
\label{fig:tile_compare}
\end{figure}

Although straightforward implementation of the super-resolution stage model on higher image resolutions achieves desired outputs, this introduces excessive requirements of the CUDA memory, which is unbearable on the resolution of $4096\times 4096$. Tiled diffusion~\cite{bar2023multidiffusion}~\cite{jimenez2023mixture} is a series of inference methods for diffusion models tackling such an issue. It separates an inference step of large images into overlapped smaller blocks and mix them together to obtain the overall prediction of the step. As shown in Figure~\ref{fig:tile_compare}, comparable results can be achieved with tiled inference. This enables CogView3 to generate images with higher resolution by a limited CUDA memory usage. It is also possible to generate $4096\times 4096$ images with tiled methods, which we leave for future work.

\vspace{-3em}
\section{Conclusion}

In this work, we propose CogView3, the first text-to-image generation system in the framework of relay diffusion. CogView3 achieves preferred generation quality with greatly reduced inference costs, largely attributed to the relaying pipeline. By iteratively implementing the super-resolution stage of CogView3, we are able to achieve high quality images of extremely high resolution as $2048\times 2048$.

Meanwhile, with the incorporation of data re-captioning and prompt expansion into the model pipeline, CogView3 achieves better performance in prompt understanding and instruction following compared to current state-of-the-art open-source text-to-image diffusion models.

We also explore the distillation of CogView3 and demonstrate its simplicity and capability attributed to the framework of relay diffusion. Utilizing the progressive distillation paradigm, the distilled variant of CogView3 reduces the inference time drastically while still preserves a comparable performance.

\newpage
\bibliographystyle{splncs04}
\bibliography{egbib}
\newpage
\appendix

\section{Sampler Derivation}\label{sec:app_sampler}

In this section, we aim to demonstrate that our designed latent relay sampler matches with the forward process of latent relay diffusion. That is, we need to prove that if the joint distribution holds,
\begin{equation}
    q(\Vec{z}_{t-1} | \Vec{z}_t,\Vec{z}_0) = \mathcal{N}(\Vec{z}_{t-1} | a_t\Vec{z}_t + b_t\Vec{z}_0 + c_t\Vec{z}_0^t, \delta_t^2\mathbf{I}),
\end{equation}
where $a_t=\sqrt{\sigma_{t-1}^2-\delta_t^2}/\sigma_t$, $b_t=1/t$, $c_t=(t-1)/t - a_t$, then 
the marginal distribution holds,
\begin{equation}
\begin{aligned}
    q(\Vec{z}_t|\Vec{z}_0)  &= \mathcal{N}(\Vec{z}_t|\Vec{z}_0^t, \sigma^2_t\mathbf{I}),\ t\in\{1,\cdots,T_r\}, \\
    \Vec{z}_0^t &= \mathcal{F}(\Vec{z}_0, t)=\frac{T_r-t}{T_r}\Vec{z}_0+\frac{t}{T_r}\Vec{z}^L. \\
\end{aligned}
\end{equation}
\textit{proof.}

Given that $q(\Vec{z}_{T_r}|\Vec{z}_0)=\mathcal{N}(\Vec{z}^L, \sigma_{T_r}^2\mathbf{I})$, we employ mathematical induction to prove it. Assuming that for any $t\leq T_r$, $q(\Vec{z}_t|\Vec{z}_0) = \mathcal{N}(\Vec{z}_0^t, \sigma^2_t\mathbf{I})$. Next we only need to prove that $q(z_{t-1}|\Vec{z}_0) = \mathcal{N}(\Vec{z}_0^{t-1}, \sigma^2_{t-1}\mathbf{I})$ holds, then it holds for all $t$ from $T_r$ to 1 according to the induction hypothesis.

First, based on
\begin{equation} \label{eq:jointly_dist}
    q(\Vec{z}_{t-1} | \Vec{z}_0) = \int q(\Vec{z}_{t-1} | \Vec{z}_t,\Vec{z}_0)q(\Vec{z}_t | \Vec{z}_0)d\Vec{z}_t,
\end{equation}
we have that 
\begin{equation}
    q(\Vec{z}_{t-1} | \Vec{z}_t,\Vec{z}_0) = \mathcal{N}(\Vec{z}_{t-1} | a_t\Vec{z}_t + b_t\Vec{z}_0 + c_t\Vec{z}_0^t, \delta_t^2\mathbf{I})
\end{equation}
and
\begin{equation}
    q(\Vec{z}_t|\Vec{z}_0) = \mathcal{N}(\Vec{z}_0^t, \sigma^2_t\mathbf{I}).
\end{equation}
Next, from Bishop and Nasrabad~\cite{bishop2006pattern}, we know that $q(\Vec{z}_{t-1} | \Vec{z}_0)$ is also Gaussian, denoted as $\mathcal{N}(\Vec{\mu}_{t-1}, \mathbf{\Sigma}_{t-1})$.
So, from Equation~\ref{eq:jointly_dist}, it can be derived that
\begin{equation}
\begin{aligned}
    \Vec{\mu}_{t-1} &= a_t\Vec{z}_0^t+b_t\Vec{z}_0+c_t\Vec{z}_0^t \\
                    &= \frac{\sqrt{\sigma_{t-1}^2-\delta_t^2}}{\sigma_t}\Vec{z}_0^t + \frac{\Vec{z}_0}{t} + (\frac{t-1}{t}-\frac{\sqrt{\sigma_{t-1}^2-\delta_t^2}}{\sigma_t})\Vec{z}_0^t \\
                    &= \frac{\Vec{z}_0^t}{t} + \frac{t-1}{t}\Vec{z}_0^t \\
                    &= \Vec{z}_0^{t-1} \quad \text{(based on Equation~\ref{eq:zt_define})}
\end{aligned}
\end{equation}
and
\begin{equation}
\begin{aligned}
    \mathbf{\Sigma}_{t-1}   &= a_t^2\sigma_t^2 + \delta_t^2 \\
                            &= (\frac{\sigma_{t-1}^2-\delta_t^2}{\sigma_t^2})\sigma_t^2 + \delta_t^2 \\
                            &= \sigma_{t-1}^2 \\
\end{aligned}
\end{equation}
In summary, $q(z_{t-1}|\Vec{z}_0) = \mathcal{N}(\Vec{z}_0^{t-1}, \sigma^2_{t-1}\mathbf{I})$. The inductive proof is complete.

\section{Supplements of Text Expansion}\label{sec:app_text}

We use the following passage as our template prompting GPT-4V to generate the grouth truth of the recaption model:

\begin{lstlisting}[language=Python, basicstyle=\tiny\ttfamily, breaklines=true,columns=flexible,showstringspaces=false]
**Objective**: **Give a highly descriptive image caption. **. As an expert, delve deep into the image with a discerning eye, leveraging rich creativity, meticulous thought. Generate a list of multi-round question-answer pairs about the image as an aid and final organise a highly descriptive caption. Image has a simple description.
**Instructions**:
- **Simple description**: Within following double braces is the description: {{<CAPTION>}}. 
  - Please note that the information in the description should be used cautiously. While it may provide valuable context such as artistic style, useful descriptive text and more, it may also contain unrelated, or even incorrect, information. Exercise discernment when interpreting the caption.
  - Proper nouns such as character's name, painting's name, artistic style should be incorporated into the caption.
  - URL, promoting info, garbled code, unrelated info, or info that relates but is not beneficial to our descriptive intention should not be incorporated into the caption.
  - If the description is misleading or not true or not related to describing the image like promoting info, url, don't incorporate that in the caption.
- **Question Criteria**:
  - **Content Relevance**: Ensure questions are closely tied to the image's content.
  - **Diverse Topics**: Ensure a wide range of question types
  - **Keen Observation**: Emphasize questions that focus on intricate details, like recognizing objects, pinpointing positions, identifying colors, counting quantities, feeling moods, analyzing description and more.
  - **Interactive Guidance**: Generate actionable or practical queries based on the image's content.
  - **Textual Analysis**: Frame questions around the interpretation or significance of textual elements in the image.
- **Note**:
  - The first question should ask for a brief or detailed description of the image.
  - Count quantities only when relevant.
  - Questions should focus on descriptive details, not background knowledge or causal events.
  - Avoid using an uncertain tone in your answers. For example, avoid words like "probably, maybe, may, could, likely".
  - You don't have to specify all possible details, you should specify those that can be specified naturally here. For instance, you don't need to count 127 stars in the sky.
  - But as long as it's natural to do so, you should try to specify as many details as possible.
  - Describe non-English textual information in its original language without translating it.
- **Answering Style**:
Answers should be comprehensive, conversational, and use complete sentences. Provide context where necessary and maintain a certain tone.
Incorporate the questions and answers into a descriptive paragraph.  Begin directly without introductory phrases like "The image showcases" "The photo captures" "The image shows" and more. For example, say "A woman is on a beach", instead of "A woman is depicted in the image".
**Output Format**:
```json
{
    "queries": [
        {
            "question": "[question text here]",
            "answer": "[answer text here]"
        },
        {
            "question": "[question text here]",
            "answer": "[answer text here]"
        }
    ],
    "result": "[highly descriptive image caption here]"
}
```
Please strictly follow the JSON format, akin to a Python dictionary with keys: "queries" and "result". Exclude specific question types from the question text.
\end{lstlisting}

In the prompt we fill \texttt{<CAPTION>} with the original caption, the prompt is used along with the input of images. On finetuning the recaption model, we use a template as:

\begin{lstlisting}[language=Python, basicstyle=\tiny\ttfamily, breaklines=true,columns=flexible,showstringspaces=false]
<IMAGE> Original caption: <OLD_CAPTION>. Can you provide a more comprehensive description of the image? <NEW_CAPTION>.
\end{lstlisting}

Figure~\ref{fig:recap_add} shows additional examples of the finalized recaption model.

\section{Details of Human Evaluation}\label{sec:app_human}

\begin{figure}[H]

  \centering
  \includegraphics[width=1.0\linewidth]{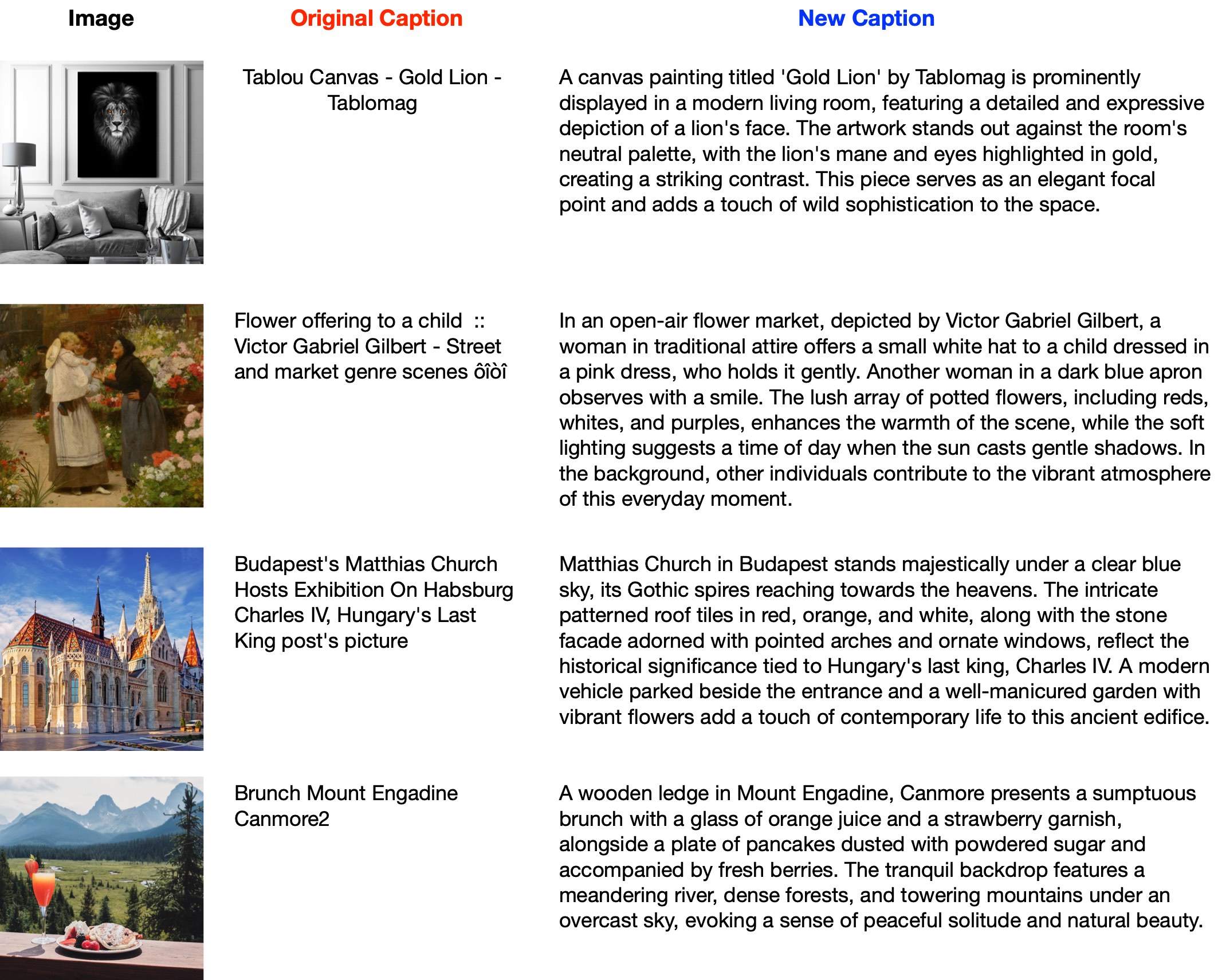}
  \caption{Examples of the recaption model results.}
  \label{fig:recap_add}

  \vspace{1cm} 

  \includegraphics[width=0.8\linewidth]{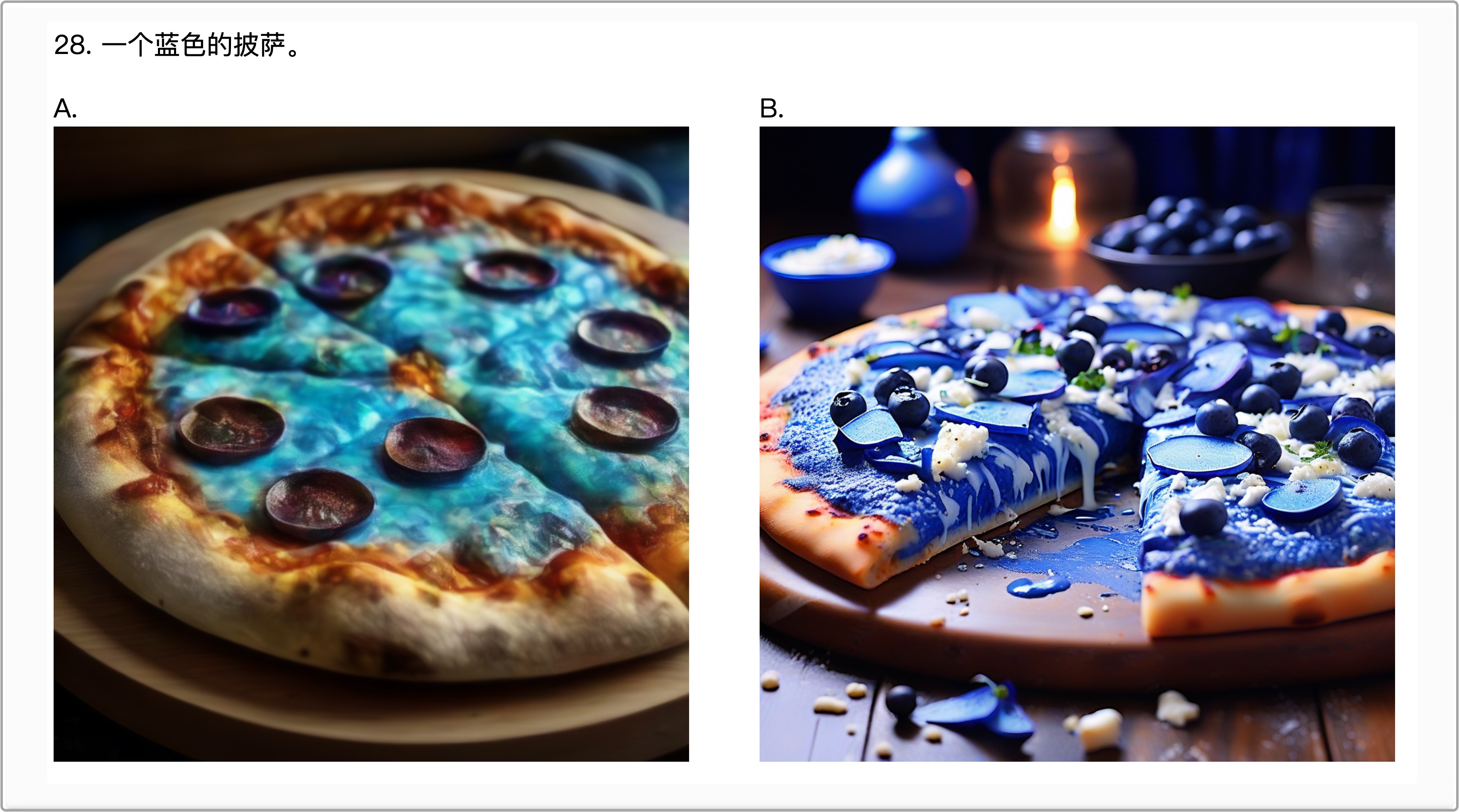}
  \caption{Interface showcases of the human evaluation. The original prompts is translated to Chinese, the mother language of our human annotators, for evaluation.}
  \label{fig:interface}
\end{figure}

Figure~\ref{fig:interface} shows a case of the interface for the human evaluation. We shuffle the order of the comparison pairs by A/B in advance and provide human annotators with equal pairs from all the comparative groups. The annotators are asked to scroll down the interface and record their preference for each pair of comparison.



\section{Additional Qualitative Comparisons}\label{sec:app_compare}

\vspace{-3em}
\subsection{Qualitative model Comparisons}

\vspace{-4em}
\begin{figure}[H]
\begin{center}
{\includegraphics[width=1\linewidth]{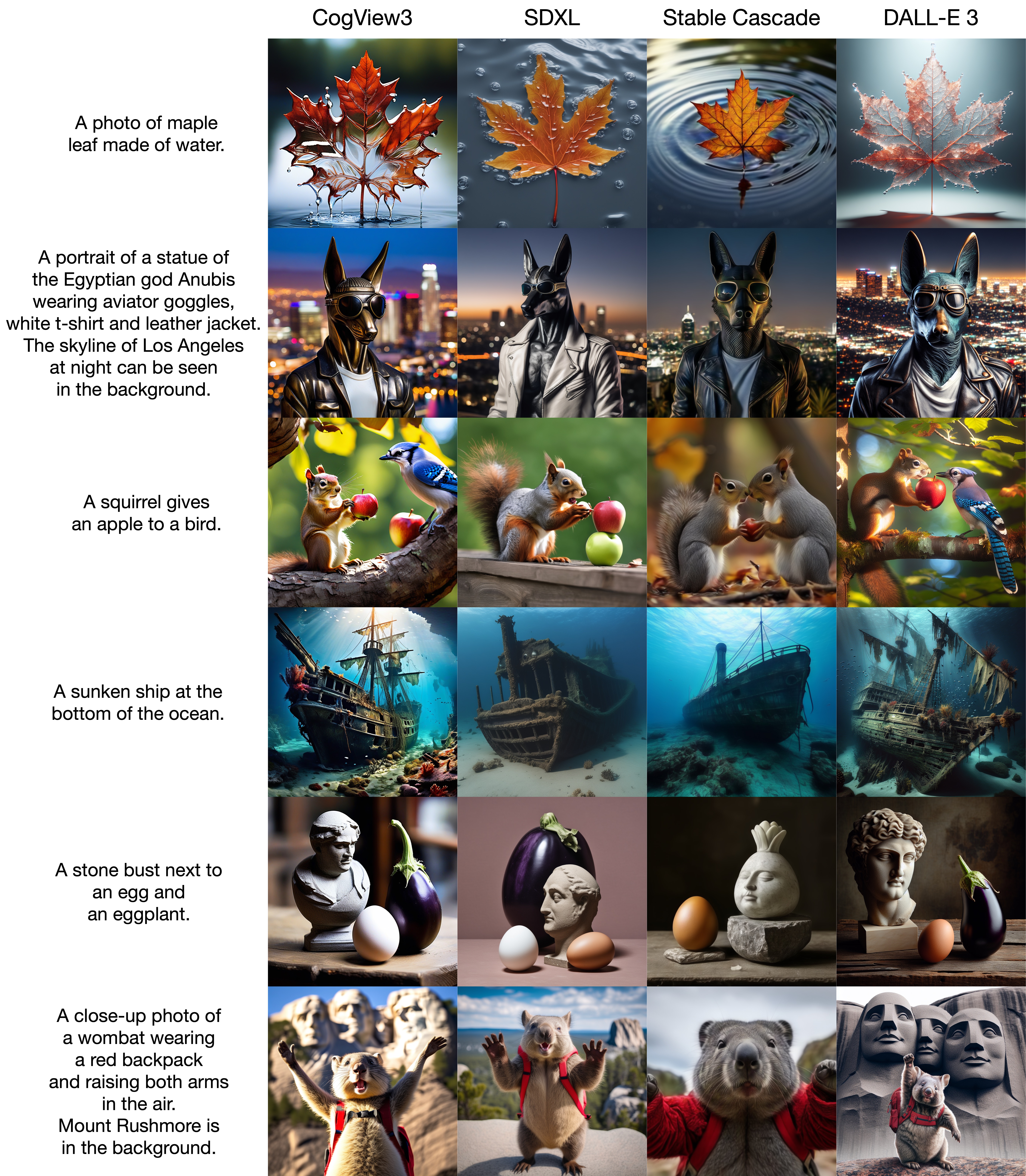}}
\end{center}
\caption{Qualitative comparisons of CogView3 with SDXL, Stable Cascade and DALL-E 3. All prompts are sampled from Partiprompts.}
\label{fig:additon_samples}
\end{figure}

\subsection{Qualitative comparisons Between Distilled Models}
\begin{minipage}{\textwidth}
\begin{figure}[H]
\centering
\includegraphics[width=1\linewidth]{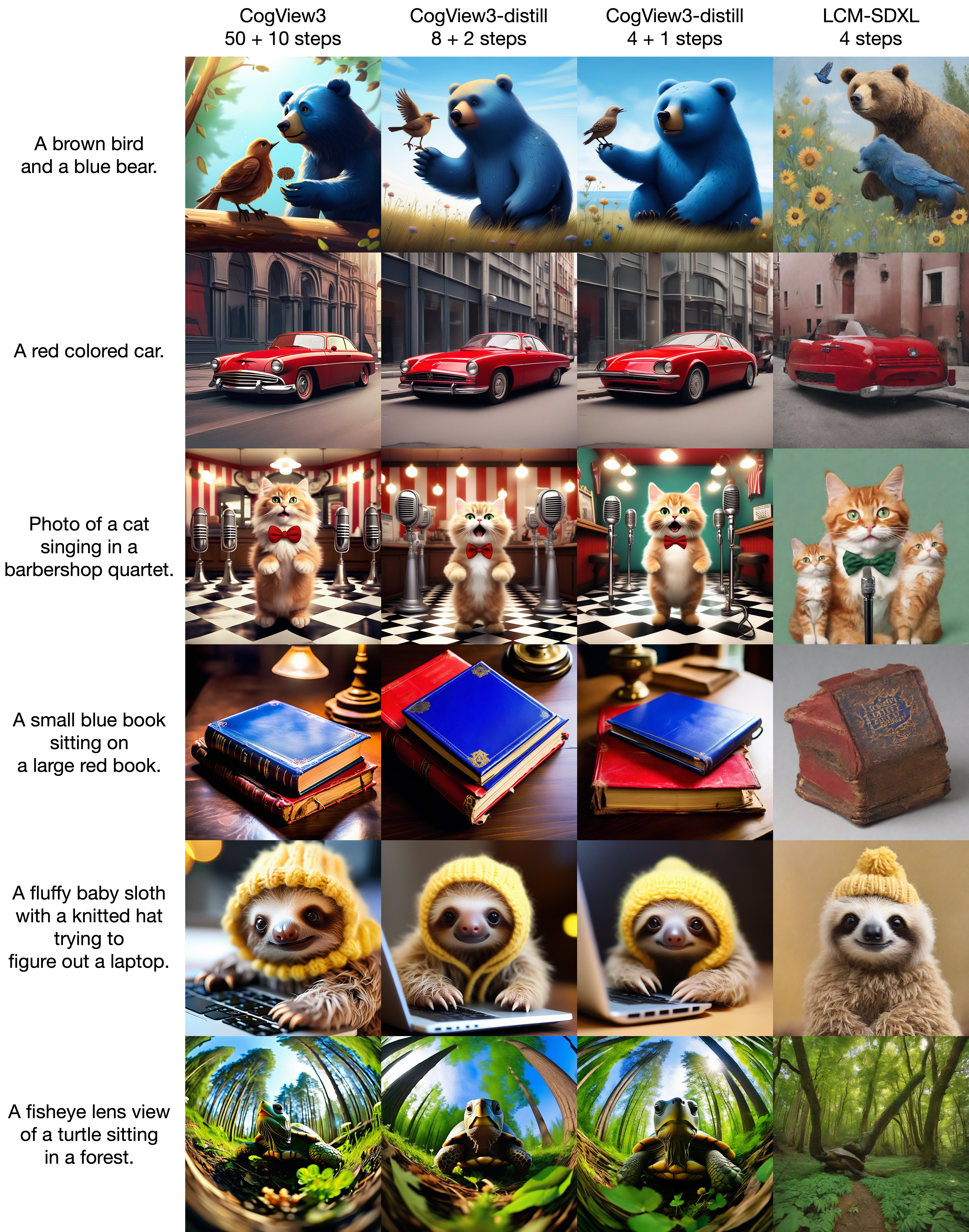}
\caption{Qualitative comparisons of CogView3-distill with LCM-SDXL, recent model of diffusion distillation capable of generating $1024 \times 1024$ samples. The first column shows samples from the original version of CogView3.}
\label{fig:additon_samples}
\end{figure}
\end{minipage}

\end{document}